%% file: ORASP-IROS.tex
\documentclass[letterpaper, 10 pt, conference]{ieeeconf}  

\IEEEoverridecommandlockouts                              

\usepackage{graphicx}
\usepackage{amsfonts}
\usepackage{amsmath}
\usepackage{amssymb}
\usepackage[font=small,labelfont=bf]{caption}
\usepackage{subcaption}
\usepackage{multirow}
\usepackage{algorithm}
\usepackage{algpseudocode}
\usepackage[normalem]{ulem}

\usepackage[dvipsnames]{xcolor}
\usepackage{cite}
\makeatletter
\let\NAT@parse\undefined
\makeatother

\DeclareMathOperator*{\argmax}{arg\,max}

\definecolor{mycitecolor}{RGB}{71, 191, 38}
\definecolor{mylinkcolor}{RGB}{40, 115, 201}

\usepackage[bookmarks=true, colorlinks, citecolor=mycitecolor,linkcolor=mylinkcolor,urlcolor=mycitecolor ]{hyperref}

\title{\LARGE \bf Optimal Robotic Assembly Sequence Planning (ORASP):\\ A Sequential Decision-Making Approach}

\author{Kartik Nagpal$^{1}$ and Negar Mehr$^{2}$
    \thanks{$^{1}$Kartik Nagpal is with the Department of Mechanical Engineering, University of California Berkeley, Berkeley, CA, 94720, USA {\tt\small kartiknagpal@berkeley.edu}}%
    \thanks{$^{2}$Negar Mehr is Faculty with the Department of Mechanical Engineering, University of California Berkeley, Berkeley, CA, 94720, USA {\tt\small negar@berkeley.edu}}%
    \thanks{This material is based upon work supported by the National Science Foundation Graduate Research Fellowship Program under Grant No. CCF-2423131 and Grant No. DGE-2146752. Any opinions, findings, and conclusions or recommendations expressed in this material are those of the author(s) and do not necessarily reflect the views of the National Science Foundation.}
}

\begin{document}
\maketitle
\thispagestyle{empty}
\pagestyle{empty}

\newcommand{\spname}{ORASP Search } 
\begin{abstract}
    The optimal robotic assembly planning problem entails determining the sequence of actions for a robot to feasibly assemble a product from its components which minimizes a given objective. This problem is made especially challenging as the number of potential sequences increase exponentially with respect to the number of parts in the assembly. Additionally, the optimal sequence must also consider and satisfy a selection of constraints such as attachment precedence or a maximum robot carry weight. Traditionally, robotic assembly planning problems have been solved using heuristics, but these methods are specific to a given robot or cost structure. In this paper, we propose to model  robotic assembly planning as a decision-making problem which enables us to use tools from shortest path algorithms and reinforcement learning. We formulate assembly sequencing as a Markov Decision Process and use Dynamic Programming~(DP) to find optimal assembly policies that far outperform the state-of-the-art in terms of speed. We further exploit the deterministic nature of assembly planning to introduce a class of optimal Graph Exploration Assembly Planners~(GEAPs) and even propose our own \spname Method. We further showcase how we can produce high-reward assembly plans for larger structures using our deep Reinforcement Learning~(RL) method. We evaluate this method on large robotic assembly problems such as the assembly of the Hubble Space Telescope, the International Space Station, and the James Webb Space Telescope. We further discuss how our DP, GEAP, and RL methods are capable of finding optimal solutions under a variety of different objective functions and how we translate any form of precedence constraints to branch pruning and further improve performance. We have published our code at \href{https://github.com/labicon/ORASP-Code}{https://github.com/labicon/ORASP-Code}.
\end{abstract}

\section{INTRODUCTION}\label{sec:introduction}
\input{1-Introduction}

\section{RELATED WORKS}\label{sec:relatedWorks}
\input{2-RelatedWorks}

\section{OUR METHOD}\label{sec:Method}
\input{3-OurMethod}

\section{EXPERIMENTAL RESULTS}\label{sec:experiments}
\input{4-ExperimentalResults}

\section{CONCLUSIONS}\label{sec:conclusions}
\input{5-Conclusions}

\section*{ACKNOWLEDGMENT}
We extend our gratitude to Dr. Robyn Woollands and the SSASSI Lab for their guidance in navigating the complexities of space applications. We would also like to thank Dr. Preston Culbertson for sharing his expertise with recreating his ILP method. Additionally, our deep appreciation goes to Dr. Saptarshi Bandyopadhyay and Dr. Amir Rahmani at JPL for their invaluable guidance and mentorship.

\bibliographystyle{ieeetr}
\bibliography{citations}  

\clearpage
\section*{APPENDIX}
\input{Appendix}

\end{document}

%% file: 1-Introduction.tex
In recent years, the surge in automated manufacturing has fueled a strong demand for assembly sequencing algorithms capable of pinpointing the best possible sequence of assembly operations~\cite{QIAN2021502}. These algorithms are essential for orchestrating robotic fleets to seamlessly integrate a myriad of parts into finished assemblies efficiently. Manufacturers have been effectively deploying these robotic assembly sequencing algorithms to greatly increase efficiency and reliability, which in turn lead to increased productivity and reduced costs~\cite{rashid2012review}. As a result, there has been immense interest in devising high performing assembly planning algorithms for these robotic workers, with numerous works in robotic furniture assembly~\cite{Knepper2013,SuarezRuiz2018}, robotic manufacturing and construction~\cite{Leder2019}, and in-space assembly~\cite{Saleh2003,Doggett}.

The assembly sequencing problem involves determining the optimal ordering of assembly operations for a fleet of robots to minimize a given objective such as production time, cost, or other relevant metrics. This robot assembly sequencing problem is made challenging by having to account for many factors, such as part compatibility or constraints made on intermediary assemblies, while ensuring optimality of the proposed solution. Traditionally, assembly sequencing problems have been solved using heuristic methods which can consistently produce near-optimal solutions. However, these heuristics exploit specific characteristics of the cost structure or dynamics, making them domain specific and incapable of generalization~\cite{Masehian2021}.

\begin{figure}
    \centering
    \hfill
    \begin{subfigure}[b]{0.42\columnwidth}
        \centering
        \includegraphics[width=0.7\textwidth]{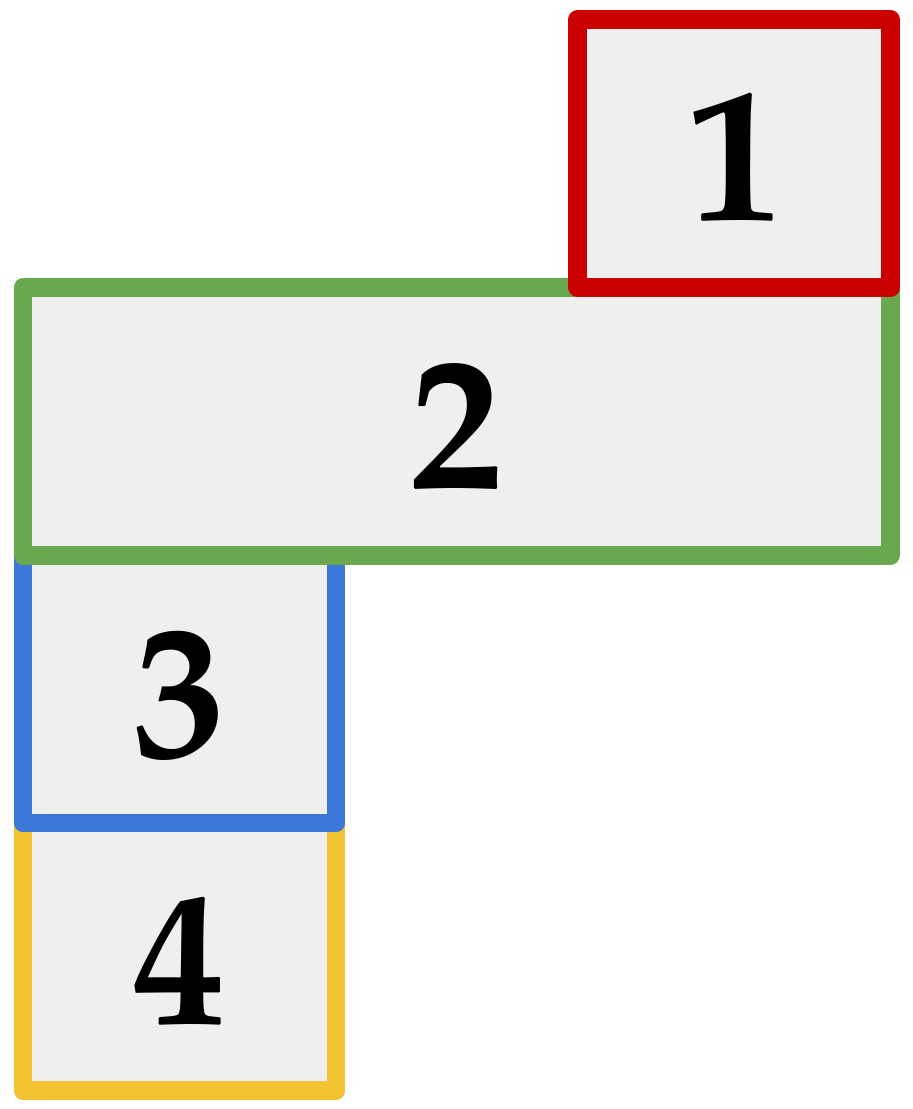}
        \caption{The simple assembly structure ``4Brick" with 4 Parts and 3 Connections.}
        \label{fig:Ex}
    \end{subfigure}
    \hfill
    \begin{subfigure}[b]{0.42\columnwidth}
        \centering
        \includegraphics[width=1\textwidth]{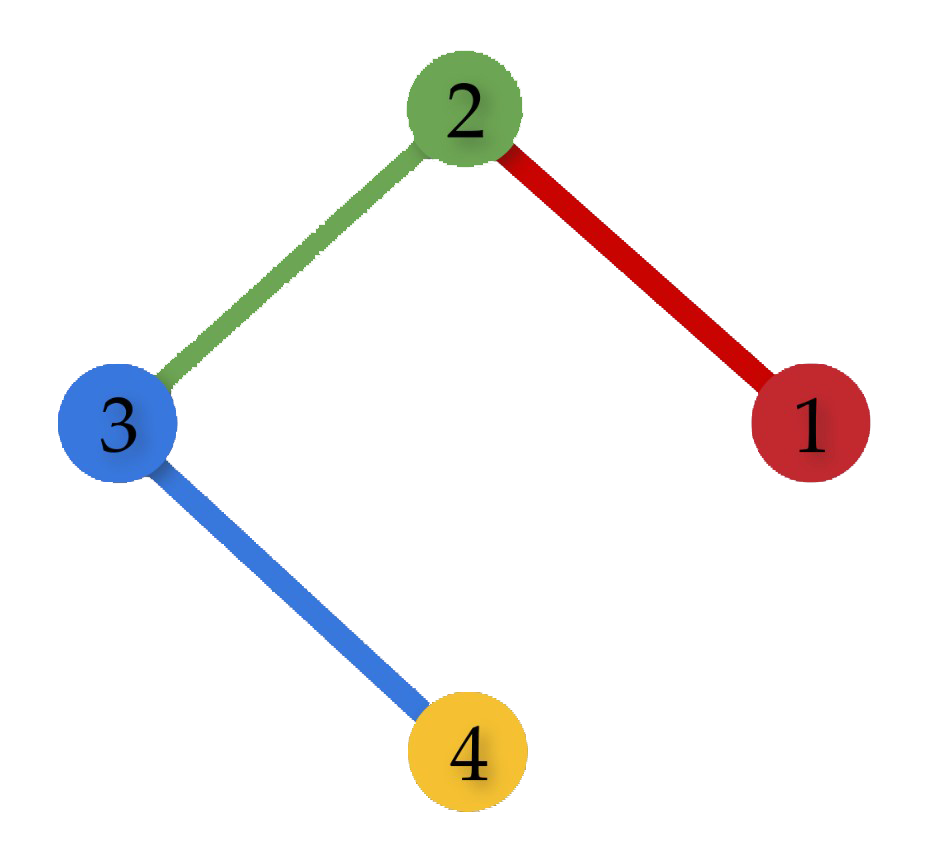}
        \caption{The graph representation $\mathcal{G}$ for the fully assembled version of the ``4Brick" Structure.}
        \label{fig:G}
    \end{subfigure}
    \hfill
    \caption{Our ``4Brick" structure of (\subref{fig:Ex}) is translated to its graph representation $\mathcal{G}$ in (\subref{fig:G}). We use this toy example repeatedly over the course of the paper.}
    \label{fig:ExampleConversion}
    \vspace{-2.1em}
\end{figure}

\begin{figure*}[!th]
  \begin{center}
      \includegraphics[width=0.95\textwidth]{figs/DQN-Results-NEW-4-Cropped.png}
      \caption{The original structures, their corresponding graph representations, and their reward distribution comparisons between {\color{OliveGreen}our DQN Approach} and a {\color{Blue}distribution of 100 randomly sampled sequences}. Our proposed DQN structure recovers the \textbf{optimal solution for the Hubble, ISS, and Furniture scenarios}. The JWST model is too large to verify for optimality with its 14 quinvigintillion potential subassemblies, but our DQN still outperforms the entire distribution of sequences. Note: Each label is followed by the tuple $(N,E)$ where $N$ is the number of parts in the structure and $E$ is the number of connections.}
      \label{fig:SpaceStructs}
  \end{center}
  \vspace{-2.25em}
\end{figure*}

Our key insight involves posing assembly sequencing as an optimal control problem with a predefined objective and set of constraints on a Markov Decision Process (MDP). We embed this problem into a novel sequential decision-making framework which finds optimal assembly plans. Our framework is effectively akin to the design of a higher level planning algorithm~\cite{Su2009,Li2022}, which is capable of optimizing for the best assembly planning sequences under a variety of objectives and sequence precedence constraints. Under our optimal control interpretation of robotic assembly planning, we define an action as a robot selecting, transporting, and attaching a piece to the current assembly and similarly define the state as the current progress of the overall assembly. To better explain our framework, we introduce a running example ``4Brick" scenario as seen in Fig.~\ref{fig:ExampleConversion}.

We further refine our optimal control interpretation to posing assembly planning as a shortest-path problem on a consolidated state-action space. A common approach used in previous papers has been to flip the assembly problem to a disassembly problem~\cite{RuzenaK.Bajcsy1989, Culbertson2019, Ghandi2015}, and have developed a group of objective functions to describe the cost of removing connections between parts as opposed to adding them. We utilize this ``Assembly-by-Disassembly'' approach in presenting our framework to align our work with these previous papers and utilize the same objective functions. Additionally, our consolidated solution space structure is symmetrical, meaning that our same approach can be equivalently explored as a forward assembly graph if there are reward or cost functions associated with these forward actions. It is also important to note that most of the heuristics discussed in our Related Works Sec.~\ref{sec:relatedWorks} are very dependent on their cost structures and cannot adapt to an ordering reversal. Using assembly-by-disassembly under our sequential decision-making framework, we optimally map from the initial state of the system (i.e. disassembled) to the final state of the system (i.e. fully assembled) under a predefined objective. We leverage tools from dynamic programming and graph traversal to introduce a class of optimal Graph Exploration Assembly Planners~(GEAPs) which provide significant improvements over the state-of-the-art~\cite{Culbertson2019} in computational runtime. Furthermore, we find that the graph upon which we calculate shortest path is our most computationally expensive step, as illustrated in Fig.~\ref{fig:GrowthComparison}. This prompted us to introduce our own \spname and Deep RL Methods which avoid the need to generate the full state-action graph.

This prompts us to further build on our formalism, and reason about assembly sequencing for very large structures using Deep Q Networks~(DQNs) and Reinforcement Learning~(RL) techniques. Using a learned assembly planning policy, we can generate near-optimal assembly sequences without having to explore the entire state-action space. Combining these practical benefits of an RL-based method with our framework allows us to adapt our approach to arbitrary assembly structures, arbitrary reward structures, and a variety of constraints on both subassemblies and the resulting assembly sequence itself. We showcase our approach on structures so large that they are intractable with the state-of-the-art, including the International Space Station~(ISS) with 32 parts and 31 connections and the James Webb Space Telescope~(JWST) with 180 parts and 256 connections. As illustrated in Fig.~\ref{fig:SpaceStructs}, our DQN approach recovers high reward earning policies for very large robotic assembly scenarios, and in fact recovers the optimal solution in the Hubble Scenario. As our approach can accommodate diverse manufacturing settings, we believe this work will help pave the way for leveraging a variety of learning-based and path-finding techniques for robot task planning.

\noindent
In summary, our main contributions are:
\begin{enumerate}
    \item A sequential decision-making framework which poses optimal assembly planning as finding optimal robotic action trajectories on a consolidated state-action space.
    \item Graph Exploration Assembly Planning~(GEAP) methods, including our \spname Method, built upon our framework for producing optimal assembly sequences for moderately-sized structures.
    \item A Deep Reinforcement Learning approach which effectively extends our framework to learning policies for the assembly planning of very large structures.
\end{enumerate}

%% file: 2-RelatedWorks.tex
As assembly planning is a multi-disciplinary problem, we explore recent advancements across a variety of different fields and discuss their respective limitations. Beginning with the biological field, assembly sequencing methods are often utilized for genomics and DNA sequencing tasks~\cite{Miller2010,Dohm2007,WarnkeSommer2016}. However, these algorithms primarily exploit the physical laws inherent to the genomics setting to produce simplified techniques like K-mer~\cite{Miller2010} and De Brujin graphs~\cite{NgDickBruijn1946}.

Meanwhile, in the manufacturing sciences, some papers employ modified Monte Carlo Tree Search~(MCTS) methods~\cite{Giorgio2018a,Chabal2022,Funk2021,Bapst2019}, but as these methods effectively operate as blind searches through the solution space, they often fail to produce optimal results. Furthermore, many of their modifications make it difficult to handle alternative reward functions or constraints during the assembly planning process. Other papers have tried to adapt Q-Learning~\cite{Giorgio2018,Hamrick2019,Kitz2021} which either rely on a tabular reward system representation or fail to ensure any kind of constraint satisfaction.

In the aerospace field, following NASA's In-Space Assembly~(ISA) operations with Hubble and the ISS, there have been numerous efforts in developing newer robotic technologies~\cite{Zhihui2021} that would enable assembling more sophisticated structures such as Asteroid Redirect Vehicles or Large Scale Space Telescopes~\cite{Belvin2016,Mukherjeea2019}. These works primarily aim to solve a coupled assembly sequencing and spacecraft motion planning problem, but in doing so are often strongly optimized for the specific dynamics models they are packaged with, and so fail to generalize well~\cite{Doerr2020,Doerr2021}. There are also other works that utilize Integer Linear Programming~(ILP) to pose assembly sequencing as a strict optimization problem. However, these methods are only capable of handling objective functions which can be expressed quadratically in the decision variables~\cite{Culbertson2019,Brown2020}. Despite its limitations, the ILP method discussed in Culbertson et al.~\cite{Culbertson2019} produces the most promising results and will be treated as our state-of-the-art solution.

Many more works aim to plan both the order of assembly and the motion guidance for the robotic arms or robots, as a combined task and motion planning problem~\cite{Lee2021,Ghasemipour2022}. However, these methods often fail to be globally optimal or are very specific to the given dynamics or cost structures. Some papers propose intricate methods for dividing the assembly into smaller structures, thus simplifying larger structures, but are rarely optimal because of this structure partitioning~\cite{Hartmann2023,Wilson1991,Agarwal2020}. Heuristic algorithms have also been built to minimize very particular objective functions~\cite{Zhao2019}. For example, the heuristic SPP-Flex~\cite{Masehian2021} focuses on minimizing locally applied stresses between parts and the Greedy and Fast algorithms in \cite{Culbertson2019} which locally minimize quadratic minimization objectives, are both very domain specific and fail to generalize.

Beyond strictly assembly planning, there have been efforts for solving generic task and motion planning problem with model-based RL methods~\cite{Bapst2019,Hamrick2019,Funk2022}. Utilizing either graph neural networks~\cite{Funk2022,Ma2022}, policy models~\cite{Bapst2019,Hamrick2019}, or sequence-to-sequence mapping~\cite{potluri2023concise} which improve sample efficiency. However, these methods largely operate on 3D coordinate state-action spaces specific to a robotic arm scenario. By operating on dense datasets they fail to generalize to long horizon tasks, such as the construction of large structures. Many of these papers also propose furniture benchmarks for evaluating their systems~\cite{Lee2021,Ghasemipour2022,Ma2022}, so we will also evaluate our method against some examples from these datasets.

%% file: 3-OurMethod.tex
\begin{figure}
    \centering
    \begin{subfigure}[b]{0.9\columnwidth}
        \centering
        \includegraphics[width=0.65\textwidth]{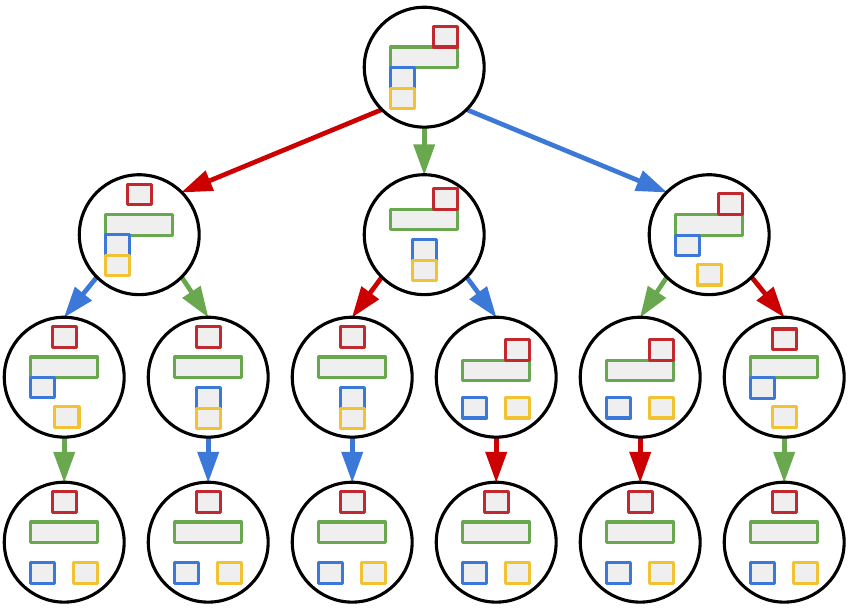}
        \caption{The state-action directed tree graph $\Tilde{\mathcal{H}}$ for the disassembly of the 4-Part, 3-Connection structure ``4Brick''.}\label{fig:treeGen}
    \end{subfigure}
    \hfill
    \begin{subfigure}[b]{0.9\columnwidth}
        \centering
        \includegraphics[width=0.55\textwidth]{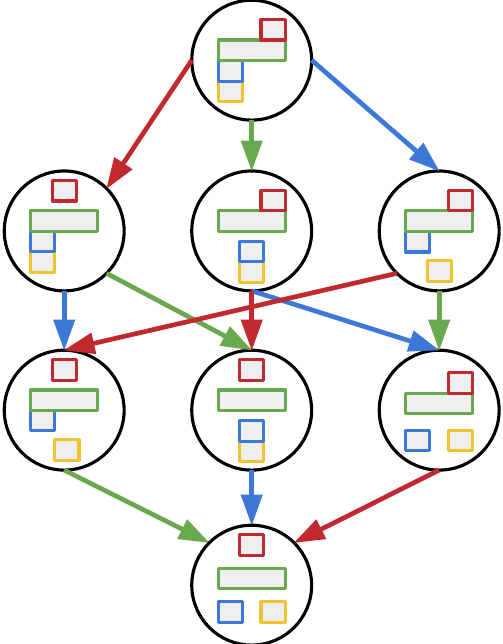}
        \caption{The \textbf{consolidated} state-action graph $\mathcal{H}$ for the disassembly of the ``4Brick'' structure.}\label{fig:treeGen2}
    \end{subfigure}
    \caption{Illustration of how we exploit the isomorphic nature of our state definition to ``consolidate'' our state-action space for the ``4Brick'' scenario. Note the color coordination between arrow colors and the connections removed at each step, which is in line with Fig.~\ref{fig:ExampleConversion}(\subref{fig:G}).}\label{fig:treeGens}
    \vspace{-1.5em}
\end{figure}

In this paper, we consider finding optimal robot assembly sequences under a pre-defined reward function. Note that when assembling a product, only the initial state (i.e. a set of unassembled pieces) and final state (i.e. the structure of the final fully-constructed product) are fixed, and our goal is to find an ordering of operations which maps the initial state to the final state. We flip the task and employ ``assembly-by-disassembly'' where the initial state $s^i$ of the system is now the fully constructed structure, and the actions $a$ are the removal of pieces or connections. Note that we primarily utilize Assembly-by-Disassembly to align our work with numerous previous works~\cite{RuzenaK.Bajcsy1989, Culbertson2019,Ghandi2015,Agarwal2020} in the assembly field, as many cost functions and general objective functions have already been engineered with this ordering in mind. We later discuss how our framework can be easily adapted to the forward assembly process as well. We refer to the fully assembled structure as the graph $\mathcal{G}$, where nodes correspond to parts in the assembly and edges represent the connections required to connect these parts together, as shown by our running example ``4Brick'' in Fig.~\ref{fig:ExampleConversion}.

We model assembly sequencing as a Markov Decision Process (MDP) defined by the tuple $\left\langle s^i, \mathcal{S}, \mathcal{A}, \mathcal{T}, R\right\rangle$. In our robotic assembly setting, $s^i$ is the deterministic initial state of the system (i.e. the fully assembled structure) as codified by the graph representation $\mathcal{G}$. For our running example ``4Brick'' we show this translation in Fig.~\ref{fig:ExampleConversion}. Meanwhile, the set $\mathcal{A}$ is the action space where $a \in \mathcal{A}$ corresponds to the act of removing a single connection from the assembly graph $\mathcal{G}$ and the state space $\mathcal{S}$ is comprised of states $s$ which denote semi-connected subassemblies which include only a subset of the edges present in the initial state $s^i$. The probability transition function $\mathcal{T}: \mathcal{S} \times \mathcal{S} \times \mathcal{A} \rightarrow \{0,1\}$ is deterministic for our setting such that $\mathcal{T}(s_{t+1} | s_t, a_t) = 1$ for the appropriate $s_{t+1}$ and $\mathcal{T}(s_{t+1} | s_t, a_t) = 0$ otherwise. Lastly, $R: \mathcal{S} \times \mathcal{A} \rightarrow \mathbb{R}$ is the transition reward function. We can utilize a variety of nonlinear reward functions. For example, we can use a minimum-time objective, where $R(s,a)$ represents the time cost of performing the given task $a$ in the given state $s$, or a similarly defined minimum-fuel objective. Our robotic assembly agent acts with a policy $\pi: \mathcal{S} \rightarrow \mathcal{A}$, generating a sequence of state-action-reward transitions.

Since there are a finite number of edges $N$ in a given full assembly $\mathcal{G}$, and we never add connections, a given disassembly trajectory is composed of a series of state-action pairs that transition from the initial fully assembled state $s^i$ to the final disassembled state $s_N$. As such, our optimal policy maps from the deterministic starting state $s^i$ to a trajectory of optimal state-action pairs $\tau = [(s^i,a_0), (s_1,a_1),..., (s_N,\cdot)]$ while maximizing the sum of rewards and reaching the final state which is the fully disassembled structure $s_N$. Thus, our goal is to find a return-maximizing policy $\pi^*$ that satisfies:
\vspace{-0.4em}
\begin{equation} \label{eq:optimal_pi}
    \pi^* \in \argmax_{\pi} \underset{\substack{a_t\sim\pi(\cdot|s_t),\\ \mathcal{T}(s_{t+1} | s_t, a_t)= 1}} {\mathbb{E}} \left[ \sum_{t=1}^{N} R(s_t,a_t) \biggl| s_0=s^i \right] \hspace{-1mm}.
\end{equation}
Under this setting, we illustrate the state-action space for the disassembly of our running example ``4Brick'' in Fig.~\ref{fig:treeGens}(\subref{fig:treeGen}). We represent this state-action space as the directed tree graph $\Tilde{\mathcal{H}}$, with the root being the fully assembled structure $s^i$, the edges corresponding to the removal of certain connections $a$, and all the leaves corresponding to the states where the structure is fully disassembled $s_N$.

Our definition of state $s$ is isomorphic to the ordering of edge removals, meaning that two states that arise from the removal of the same connections but in alternative orders are effectively identical, and can therefore be combined. Hence, we combine identical nodes on the state-action space graph $\Tilde{\mathcal{H}}$, and produce the ``consolidated'' graph $\mathcal{H}$ shown on Fig.~\ref{fig:treeGens}(\subref{fig:treeGen2}). Consequently, the growth rate of the state-action space is greatly reduced, which makes much larger structures now computationally tractable, as illustrated in Fig.~\ref{fig:GrowthComparison}. Our state-action consolidation technique provides similar benefits to those seen in older assembly planning papers using AND/OR graphs~\cite{Mello1991,Wilson1991,Zhao2019}, without as much of a computational overhead. Additionally, our consolidated graph structure is symmetrical, meaning that our generated graph can be equivalently explored as a forward assembly graph with just a small adjustment to the reward function definition. In some cases, the costs of the forward assembly and backward disassembly tasks are the same, meaning that no change in weights is necessary, and our method can be adapted immediately. Now, with our consolidated state-action space $\mathcal{H}$, we effectively reduce the assembly planning problem to a path-finding problem from the start node $s^i$, to the end node $s_N$ with the reward structure encoded as edge weights. Hence, we just need to find the optimal path and reverse the list of actions taken to produce the optimal assembly sequence.

\subsection{Dynamic Programming}
In this section we will establish three different methods to obtain optimal assembly sequences. To begin, we simply adapt the traditional dynamic programming approach to explore our consolidated state-action graph and recover the optimal path and thus the optimal assembly sequence. Let $V_k(s): \mathcal{S} \rightarrow \mathbb{R}$ denote the $k$-th iteration of a value function which maps our description of a partially constructed assembly $s$ to the value of being in that state. Akin to standard dynamic programming, we initialize $V_0(s) = 0$ for all $s$ and iteratively apply the Bellman update
\vspace{-0.8em}
\begin{equation}\label{eq:value}
    V_{k+1}(s) \gets \max_{a\, \in\, \mathcal{H}[s]} \big\{ R(s,a) + V_k(\mathcal{H}[s][a]) \big\},
\end{equation}
where $\mathcal{H}[s]$ denotes the connections that can be feasibly removed for our current partial assembly $s$ as recovered by enumerating the edges on $\mathcal{H}$ that lead away from the node $s$. Similarly, $\mathcal{H}[s][a]$ denotes the deterministic next state after taking edge $a$ from node $s$. As the state-action space is finite, our value iteration algorithm converges in a finite number of steps. Using this optimal value function $V^*$, we roll out the optimal policy $\pi^*(a|s)=\argmax_{a \in \mathcal{H}[s]} \left[R(s,a)+V^*(\mathcal{H}[s][a])\right]$. This allows us to directly recover the optimal trajectory of state-action pairs and thus the optimal assembly sequence.

\subsection{Graph Exploration Assembly Planners~(GEAPs).} Continuing, we can further exploit the deterministic nature of our motivating robotic manufacturing setting and the structure of our consolidated state-action graph $\mathcal{H}$. Because of these deterministic dynamics, the shortest path from the initial state to the final state is the optimal assembly strategy. Additionally, for suitably small structures, we can fully generate our graph $\mathcal{H}$, which means we can deploy any optimal shortest-path algorithm to fully explore the graph and recover the optimal solution. We refer to these optimal path-finding algorithms as Graph Exploration Assembly Planners~(GEAPs). As such, we can now harness previous works on optimal shortest path and graph traversal algorithms~\cite{Cormen2022} to optimally solve assembly sequencing problems. Further note that on directed graphs there is no need for re-visitation, meaning that search algorithms like Breath-First Search~(BFS) can fully explore the graph with time complexity $\mathcal{O}(N)$ where $N$ is the number of vertices in $\mathcal{H}$. As such, GEAP Algorithms like Dijkstra's (which adapts BFS to a shortest path algorithm) and Bellman-Ford shortest path can produce optimal solutions at radical speeds. 

\subsection{\spname Method.} However, generating the full state-action graph $\mathcal{H}$ can be very computationally taxing, and so we propose our own \spname Method which only generates portions of the graph as necessary. Assume that our reward function is non-positive, i.e., $R(s,a) \leq 0$ for all $s$, $a$. This is a common assumption as most reward functions for assembly are either based on fuel or energy expenditure~\cite{Culbertson2019,RuzenaK.Bajcsy1989} or can be easily transformed to $\mathbb{R}^-$. Without loss of generality, let $\bar{R}$ denote the total reward acquired by an arbitrary full assembly sequence $\bar{\tau}$. For example, we can set $\bar{R}$ as the total reward of a sequence we achieve by making random choices at each step. Then, as we generate our consolidated graph $\mathcal{H}$, if we incur an intermediary running reward $R_t < \bar{R}$, then there is no manner in which the total reward of this branch will outperform $\bar{\tau}$. This means that we can immediately discard the branch as suboptimal. Using this intuition, we propose our \spname Method which generates the graph as it explores. \spname initially samples a random disassembly sequence, defines $\bar{R}$ and then recursively expands through the state-action space, generating the consolidated state-action graph $\mathcal{H}$ in the process, and if a given branch produces a running reward worse than $\bar{R}$, it is abandoned. Additionally, whenever \spname finds a complete disassembly sequence that outperforms $\bar{R}$, we update the value of $\bar{R}$ itself and store $\bar{\tau}$. We continue exploring in this recursive manner until all threads of the process are completed. As a result, we can find the optimal sequence $\tau^*$ without generating large portions of the graph.

When we combine these low time-complexity GEAPs with our state-action consolidation technique, our approaches provide orders of magnitude faster runtimes for the optimal assembly planning problem and easily handle moderately sized structures, as discussed in Sec.~\ref{sec:experiments}.

\begin{figure}
    \centering
    \includegraphics[width=0.95\columnwidth]{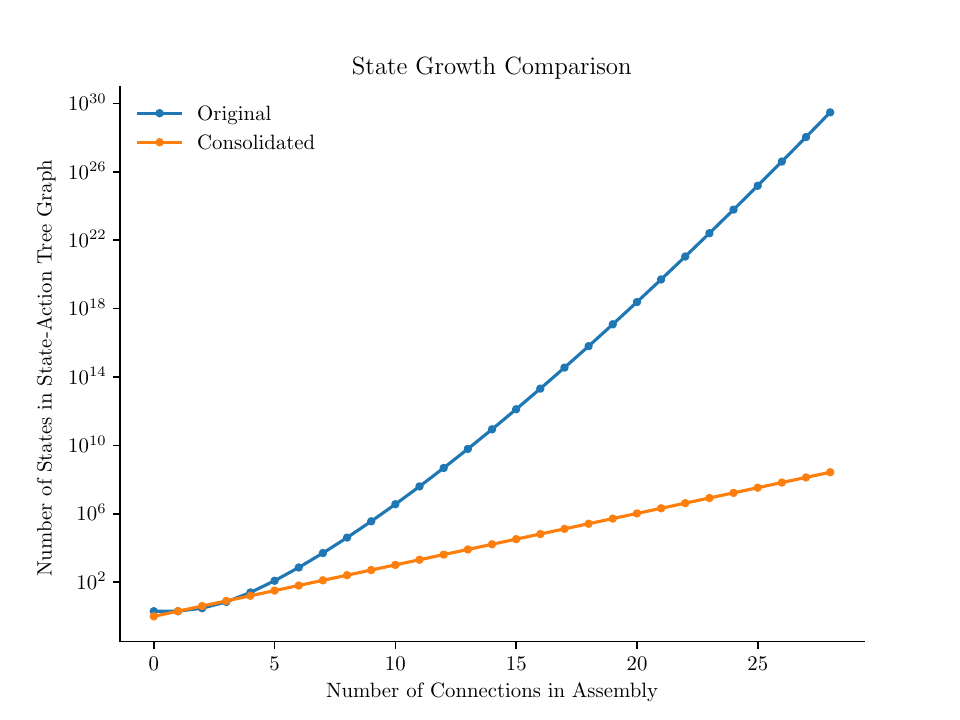}
    \caption{The rate at which the number of states grows compared to the number of connections in the fully assembled structure. This is equivalent to the growth rate of the number of nodes in the consolidated graph $\mathcal{H}$. Note the y-axis of these plots are on a logarithmic scale.}\label{fig:GrowthComparison}
    \vspace{-2em}
\end{figure}

\subsection{Our Deep Q Network Method}
Despite these improvements, the growth rate of the state and action space is exponential with respect to the number of parts in the assembly, as illustrated in Fig.~\ref{fig:GrowthComparison}. This observation prompts an extension of our framework to harness a deep Reinforcement Learning~(RL) based approach for finding high reward earning assembly sequencing policies. We propose to utilize a Deep Q Network~(DQN) which actively learns and explores the state-action space of the assembly generation problem. We draw from similar papers of the RL community which demonstrate the effective use of RL policy gradient methods for solving vehicle routing and path-finding problems~\cite{Kool2018,Czech2020,Silver2017}. For context, a DQN is a RL algorithm that uses a deep Artificial Neural Network~(ANN) as a universal function approximator for the optimal quality function $Q^*: \mathcal{S} \times \mathcal{A} \rightarrow \mathbb{R}$ which associates a numerical ``quality'' to the action of removing a given connection $a$ when operating on the given assembly state $s$. This deep ANN structure allows the DQN to handle higher-dimensional input spaces and to learn directly from state-action-reward data.

Since the DQN is based on this deep ANN structure, our graphical representation of state must be translated into a compliant form. To achieve this, we propose a simple indicator function to represent a given state $s$ as a vector $s \in \mathbb{R}^N$ where $N$ is the number of edges in the completed assembly, and each element $s_j = 1$, $j \in \{0,1,\ldots,N\}$ if the corresponding edge $E_j$ in the graphical representation of state $s$ exists (i.e. there is a connection), and $s_j=0$ otherwise. As such, the initial fully assembled state $s^i$ is represented by the one vector $\mathbf{1}_N=(1,1,\ldots,1) \in \mathbb{R}^N$ and the fully deconstructed state $s_N$ is the zero vector $\mathbf{0}_N$. The output of this DQN is the vector $q \in \mathbb{R}^N$ with $q_j$ estimating the Q-value of the action of removing connection $E_j$.

During the DQN training, we utilize an $\varepsilon$-greedy policy~\cite{Stadie2015} to ensure that the trajectories fed to the ANN are sufficiently exploratory. Our implementation utilizes a decay for the $\varepsilon$ term, wherein later episodes' trajectories are more based on our DQN estimate of the quality function rather than random actions. Therefore, our algorithm explores more the parts of the solution space producing better results. Furthermore, we introduce a very simple curriculum~\cite{soviany2022curriculum} where each episode begins with only a small number of connections left and the DQN is only responsible for finishing the disassembly task. Over the course of training, the number of connections in this initial state is iteratively increased until the DQN is performing the entire disassembly process in every episode. Lastly, the DQN employs an experience replay buffer during training to reduce correlations between consecutive updates of the network and thus improve sample efficiency~\cite{Fedus2020}. 

\subsection{Constraints} \label{sec:Constraints}
In this section, we discuss the versatility of our formulation in accommodating various constraints. Our framework not only guarantees satisfaction but also uses these constraints to significantly reduce computational runtimes. We primarily focus on sequential precedence constraints which codify when one piece is required to be removed before another piece is removed. For instance, the center part in the lattice structure, as illustrated in \ref{fig:SmallStructs}, must be placed before the parts around it are placed. We handle this constraint by setting the probability transition function $\mathcal{T}(s_{t+1}|s_t,a_t) = 0$ for state transfers that would violate our constraints. Alternatively, if we are aware of the constraints before generating our consolidated state-action graph $\mathcal{H}$, then we can prune or abandon any newly generated branches which violate our constraint. Both of these approaches directly ensure constraint satisfaction by construction, while also reducing the size of the state-action space and further \emph{reduce} the runtime.

Both our transition function modification and pruning approaches for handling sequential constraints help to highlight a key capability of our method: fast re-planning. Any part loss or connection inaccessibility can be easily posed as an infeasibility in the state-action space. In fact, the Bellman-Ford algorithm pre-computes the minimum distances for all point-to-point traversals, and so can re-plan around this new infeasibility constraint immediately. Similarly, our DQN can output a new sequence accounting for the infeasibility by construction. Furthermore, we can use simple re-weighting on our generated graph to find the optimal assembly sequence for a variety of nonlinear objective functions (as codified by the reward function) and constraint specifications, without ever needing to regenerate our state-action graph. This result is profound as re-weighting our generated graph directly translates to real-time optimal re-planning of the assembly sequence.

Our framework also allows for constraints which specifies both the maximum number of unconnected pieces allowed during construction as well as the maximum size of each of these pieces. We translate these Unconnected Parts Constraints~(UPCs) to our framework as a maximum on the number of isolated subgraphs $M_{num}$ allowed in the graph representation of the current state $s$, and to a maximum on the number of nodes allowed in the newly formed smaller subgraph $M_{size}$, respectively. We handle these UPCs by adding additional metadata on the size and weight of each piece in the structure to our definition of state while also storing the number of deconstruction zones as $M_{num}$ and the maximum load each robot can transport as $M_{size}$. We then evaluate actions during the search process and prune branches which would cause violations of these constraints. This way our framework can be used to simulate the transport of multi-part piece during the assembly process and harness UPCs to ensure that the resulting assembly plan is physically possible. 

Note that for both sequential constraints and UPCs, satisfaction is ensured by construction and since the DQN method directly returns Q-values, our transition function modification approach can still be used to enforce these constraints. Hence, our framework captures a multitude of constraint definitions and even utilizes these constraints to shrink the state-action space, further improving the runtime of our methods.

%% file: 4-ExperimentalResults.tex
\begin{figure}
    \centering
    \subfloat[4Brick~(4,3)]{\includegraphics[width=0.25\columnwidth]{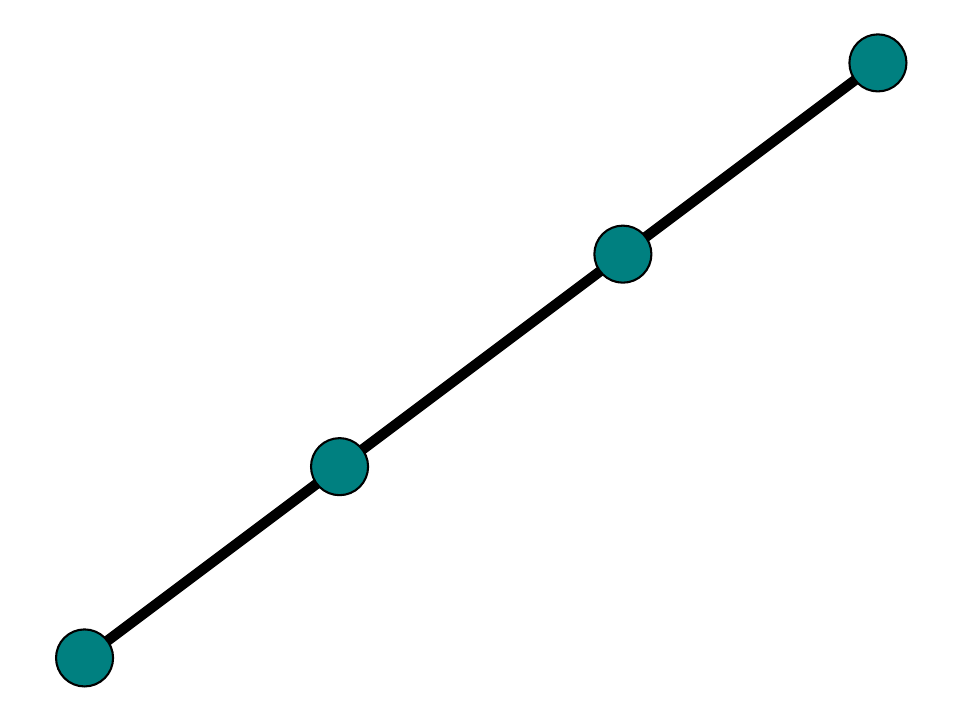}}\hfill
    \subfloat[2x3~(6,7)]{\includegraphics[width=0.25\columnwidth]{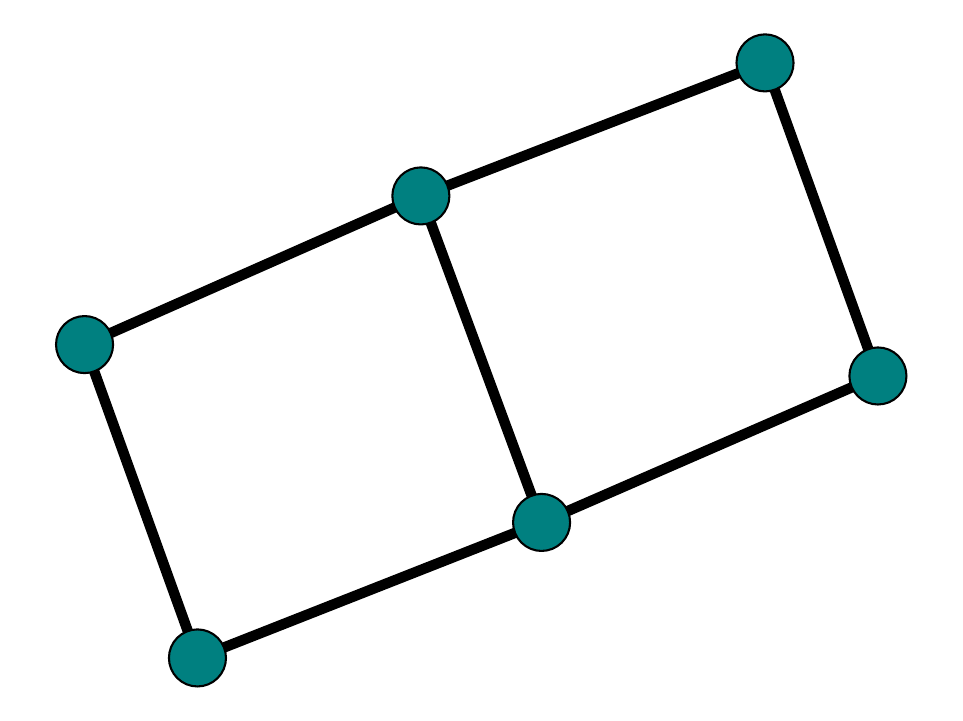}}\hfill
    \subfloat[Lattice~(9,12)]{\includegraphics[width=0.25\columnwidth]{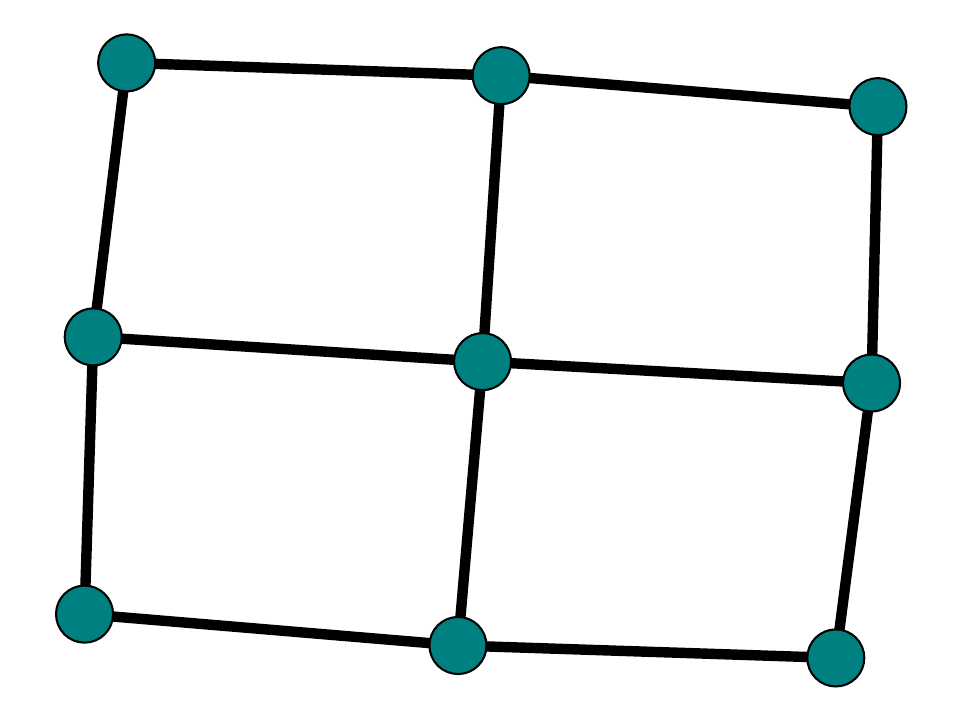}}\hfill\label{subfig:Lattice}
    \subfloat[Hubble~(20,19)]{\includegraphics[width=0.45\columnwidth]{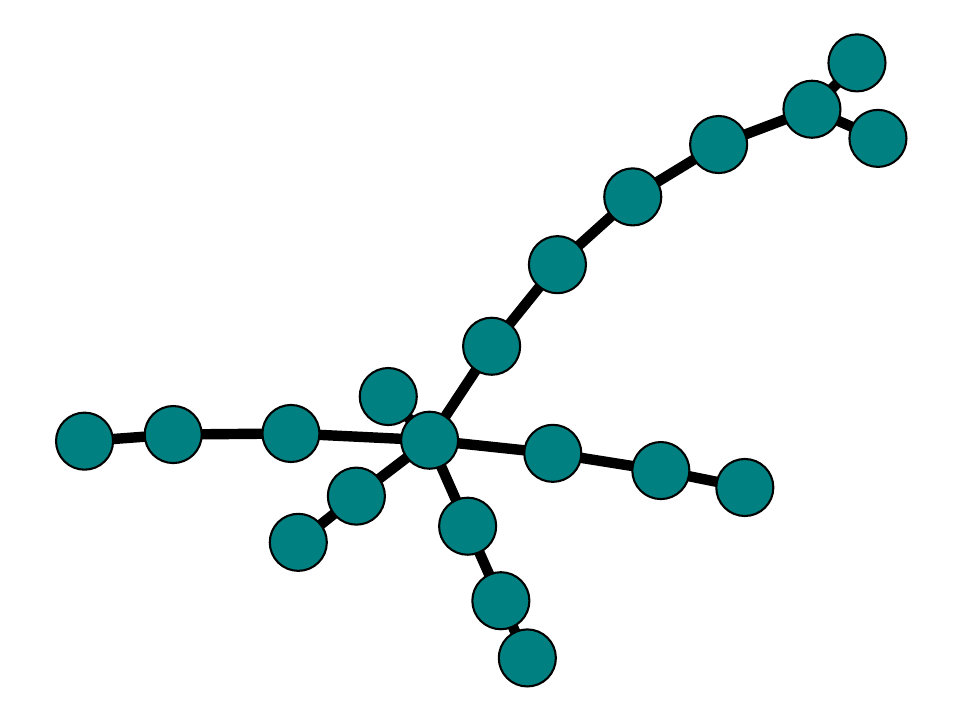}}\hfill
    \subfloat[IKEA Table~(9,8)]{\includegraphics[width=0.45\columnwidth]{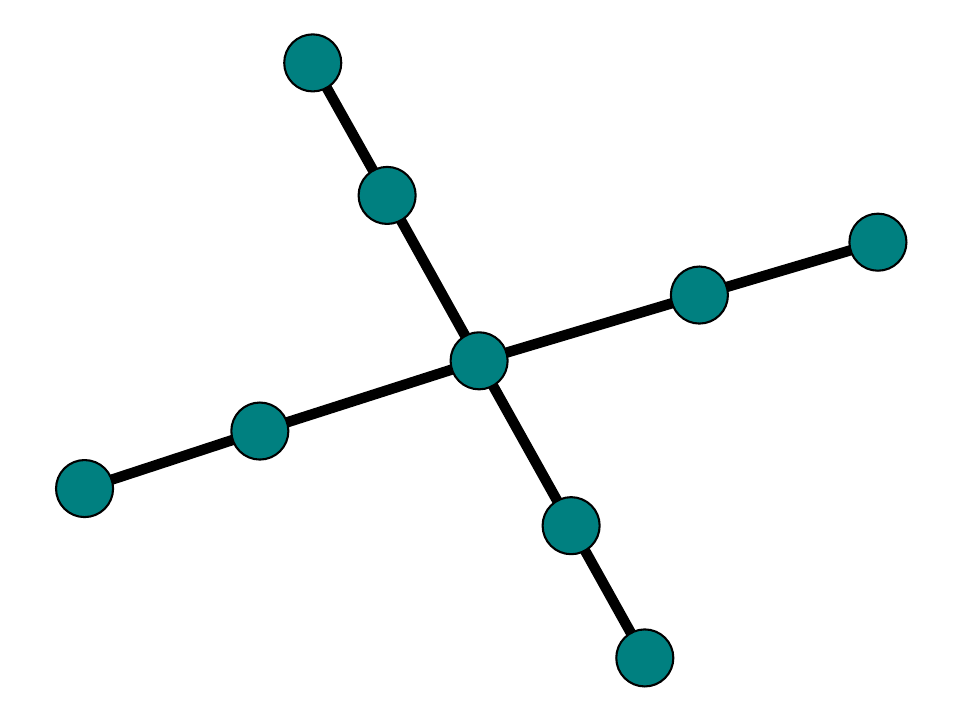}}
    \caption{Graph representations for the structures discussed in Table~\ref{table:res}. Each label is followed by the tuple $(N,E)$ where $N$ is the number of parts in the structure and $E$ is the number of connections in the structure.}\label{fig:SmallStructs}
    \vspace{-2.15em}
\end{figure}

\def\arraystretch{1}
\begin{table*}
    \centering
    \begin{tabular}{|llllll|}
        \hline
        \multicolumn{1}{|c|}{\multirow{2}{*}{\begin{tabular}[c]{@{}c@{}}\emph{All Times are}\\ \emph{in Seconds}\end{tabular}}} & \multicolumn{1}{c|}{\multirow{2}{*}{\textbf{\begin{tabular}[c]{@{}c@{}}State of the Art\\ (ILP)~\cite{Culbertson2019}\end{tabular}}}} & \multicolumn{4}{c|}{\textbf{Our Graph Exploration Assembly Planners~(GEAPs)*}} \\ \cline{3-6}
        \multicolumn{1}{|c|}{} & \multicolumn{1}{c|}{} & \multicolumn{1}{l|}{\textbf{\spname}} & \multicolumn{1}{l|}{\textbf{Value Iteration}} & \multicolumn{1}{l|}{\textbf{Dijkstra's}} & \textbf{Bellman Ford} \\ \hline
        \multicolumn{1}{|l|}{\textbf{4Brick}} & \multicolumn{1}{l|}{Min-time: 0.46} & \multicolumn{1}{l|}{\textbf{0.001000}} & \multicolumn{1}{l|}{0.002939} & \multicolumn{1}{l|}{0.001997} & 0.001997 \\ \hline
        \multicolumn{1}{|l|}{\textbf{2x3}} & \multicolumn{1}{l|}{Min-time: 1.32} & \multicolumn{1}{l|}{\textbf{0.023030}} & \multicolumn{1}{l|}{0.070786} & \multicolumn{1}{l|}{0.069553} & 0.070827 \\ \hline
        \multicolumn{1}{|l|}{\textbf{Lattice}} & \multicolumn{1}{l|}{\begin{tabular}[c]{@{}l@{}}Min-time: 4\\ Min-travel: 5800\end{tabular}} & \multicolumn{1}{l|}{\textbf{0.472041}} & \multicolumn{1}{l|}{1.354401} & \multicolumn{1}{l|}{1.322265} & 1.345777 \\ \hline
        \multicolumn{1}{|l|}{\textbf{Hubble}} & \multicolumn{1}{l|}{\begin{tabular}[c]{@{}l@{}}Min-time: 5000 \\ Min-travel: 20000\end{tabular}} & \multicolumn{1}{l|}{\textbf{6.284114}} & \multicolumn{1}{l|}{108.454639} & \multicolumn{1}{l|}{107.847125} & 107.877225 \\ \hline
        \multicolumn{1}{|l|}{\textbf{IKEA Table}} & \multicolumn{1}{l|}{} & \multicolumn{1}{l|}{\textbf{0.024000}} & \multicolumn{1}{l|}{0.143997} & \multicolumn{1}{l|}{0.139000} & 0.138998 \\ \hline
        \multicolumn{6}{|c|}{*For our methods, minimum-time and minimum-travel take the same runtime.} \\ \hline
    \end{tabular}
    \caption{Comparison of runtimes between our methods and the state-of-the-art ILP method used in Culbertson et al.~\cite{Culbertson2019}. Note that the majority of runtime for the GEAP methods is due to the graph generation times and so often have similar times. We test our methods against both the minimum-time and minimum-distance cost structures in the state-of-the-art paper. Similarly, all constraints in the state-of-the-art were translated to sequential precedence and number of subassembly constraints when possible. Note: The ``IKEA Table'' scenario uses the reward function from the Blocks Assemble! paper~\cite{Ghasemipour2022} and can not be written quadratically in terms of the decision variables and is thus impossible for the state-of-the-art. The ILP deemed ``4Brick'' and ``2x3'' under the minimum-travel reward infeasible.}
    \label{table:res}
    \vspace{-2em}
\end{table*}

In this section, we showcase both the performance of our Graph Exploration Assembly Planners~(GEAPs) against the state-of-the art on moderately-sized structures ($\leq 25$ connections). We then showcase our DQN method on large scale problems which are infeasible for the state-of-the-art. All our timed experiments were run in Python on an AMD Ryzen 9 5900X with 16~GBs of memory, while the state-of-the-art was run in Julia using JuMP~\cite{Lubin2023} and the IBM CPLEX solver on a faster Google Cloud virtual machine with 8 vCPUs and 30~GBs of memory. We attempted to run the state-of-the-art on the same hardware, but experienced worse results than those reported, and so included data from the original paper whenever available. A majority of our experiment cases are In-Space Assembly~(ISA) problems with minimum fuel or time objectives to match the experiments in the state-of-the-art paper~\cite{Culbertson2019}. We have further adapted the ``IKEA Table" and ``Furniture" scenarios from the IKEA~\cite{Lee2021} and Blocks assemble~\cite{Ghasemipour2022} benchmarks, respectively. We use a completion objective function for these scenarios to properly emulate the reward functions used in the original benchmarks.

For the GEAPs cases, we use the same reward functions as those utilized in our state-of-the-art paper~\cite{Culbertson2019} and only apply a small augmentation which prevents infeasible motions. Accordingly, the effects of the geometric and movement constraints from the state-of-the-art paper~\cite{Culbertson2019} are captured without sacrificing the runtime of our GEAP methods. Additionally, we translate the state-of-the-art's connectivity constraints into a selection of sequential constraints, such as ensuring that the outer connections of the Lattice structure must be removed before the internal connections with the central part can be removed. These translations ensure that both methods are effectively solving the same optimal control problem.

\textbf{Graph Exploration Assembly Planners~(GEAPs).}
We illustrate the power of our GEAPs on the simple structures shown in Fig.~\ref{fig:SmallStructs}. Both the state-of-the-art and our GEAP methods recover optimal solutions. As Table~\ref{table:res} demonstrates, our GEAP methods converge to the optimal solutions at least 3 times faster on the Lattice scenario and nearly 50 times faster on the Hubble scenario, all while still satisfying the same constraints imposed on the ILP method. Additionally, our graph exploration Dijkstra method produces over a 4,000 time speedup compared to the state-of-the-art for the Lattice min-travel scenario. In fact, with the exponential growth rate of the state-action space, our graph generation time is our limiting factor much more than the time complexity of the GEAPs. Comparing amongst our methods, the value iteration method has the added capability of handling stochastic behavior, but as this setting is deterministic, it is surpassed in speed and efficiency by the other two GEAPs. Among those, Dijkstra's is faster, but Bellman-Ford finds the shortest path from the given start node $s^i$ to every other node in the graph, making it much more suitable for re-planning. Our novel \spname method further speeds up the process by avoiding the need to generate large portions of the state-action tree graph. In summary, our GEAP methods provide results well surpassing the state-of-the-art and each have their own further benefits.

\textbf{Reinforcement Learning Methods.}
For the DQN cases, we utilize a highly nonlinear cost function to replicate the minimum fuel objective of a CubeSat Robotic Agent with a simple low thrust engine. As this cost function cannot be expressed quadratically in terms of the problem parameters, the state-of-the-art~\cite{Culbertson2019} is entirely incapable of handling the scenario. Furthermore, we use our unconnected parts constraints~(UPCs) to simulate the agent moving multi-part structures of maximum size $M_{size}=3$ per robotic agent and with a maximum number of construction sites $M_{num}=4$.

Along with these added cost function and constraint complexities, we also showcase our DQN method on much larger scenarios. As such, we deployed the DQN on the Hubble structure from Fig.~\ref{fig:SmallStructs} and Table Scenario~\ref{table:res} to compare it with our GEAP methods, as well as three structures too large to be computationally infeasible under the state-of-the-art. Note that due to the size of these three largest structures there are no suitable baselines for comparison and so we compare the cumulative reward of the assembly plan generated by our DQN method against a distribution of 100 randomly sampled trajectories. These random trajectories are acquired by uniformly sampling the valid action space at each step until the fully disassembled state $s_N$ is reached. We then take these 100 trajectories, calculate the cumulative rewards over each trajectory, and report the mean of the data as well as a violin-plot showing the distribution and its bounds, as seen in Fig.~\ref{fig:SpaceStructs}. For simplicity, both the Hubble and ISS scenarios were trained for 5,000 iterations using a simple PyTorch-CUDA setup on a single NVIDIA 3080~Ti and took approximately 15 minutes to train. The JWST scenario was trained for 20,000 iterations and took approximately 35 minutes. As reflected in Fig.~\ref{fig:SpaceStructs}, our DQN approach outperforms the full distributions of 100 random trajectories on every scenario. Additionally, after taking the time to run the Hubble, ISS, and Furniture scenarios with our GEAP methods under this minimum-fuel CubeSat robot reward function, we verified that our DQN approach recovers the optimal disassembly sequence for all three of those scenarios. This is particularly notable, as \emph{our DQN method takes less time to train and run than the state-of-the-art in the Hubble case, while still reproducing the optimal result}. For the JWST scenario, the assembly is too large to run with the GEAPs, and so we cannot easily verify the optimality of the discovered solution, but our DQN again outperforms the entire distribution of random sequences.

%% file: 5-Conclusions.tex
This paper presented a novel approach to the robot assembly sequencing problem by reformulating it as an optimal control problem, which was then further distilled to a shortest-path problem. We then proposed a sequential decision-making framework encompassing tools from graph theory, path-finding algorithms, and deep reinforcement learning to far surpass the state-of-the-art, both in terms of efficiency and scalability. Our framework also enabled us to handle assembly scenarios orders of magnitude larger and under a variety of nonlinear reward functions, sequential precedence constraints, and number-of-subassemblies constraints.

\textbf{Limitations and Future Work.}
Since the state-action space grows exponentially with the number of connections in the assembly structure, our graph exploration assembly planning methods eventually become computationally intractable. Our DQN approach largely mitigates this limitation, but itself introduces a potential lack of global optimality. We mostly considered deterministic test cases, but our sequential decision making framework can be easily extended to simulate uncertainties, such as chances of failure. Additionally, if our method is paired with existing systems for resource tracking, agent assignment, and motion planning, it would provide a full-stack solution to the robotic manufacturing problem.

%% file: Appendix.tex
In this section, we provide additional results and supplementary materials to help support the claims made in the body of the paper.

\subsection*{State and Action Growth}\label{sec:StateActionGrowth}
\begin{figure}
    \centering
    \subfloat[State Growth Rate Graph]{\includegraphics[width=0.95\columnwidth]{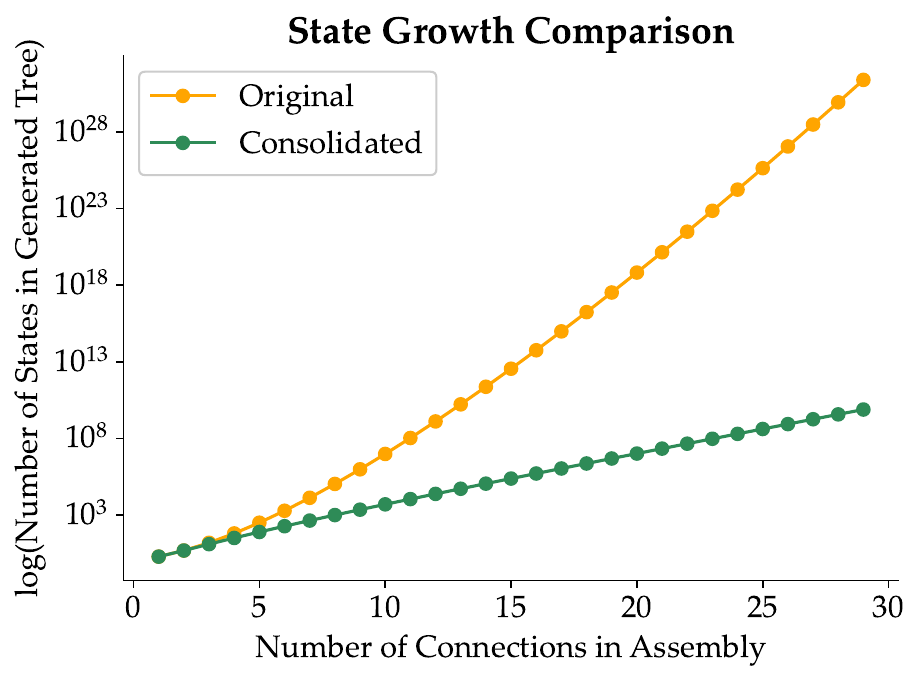}}

    \subfloat[Action Growth Rate Graph]{\includegraphics[width=0.95\columnwidth]{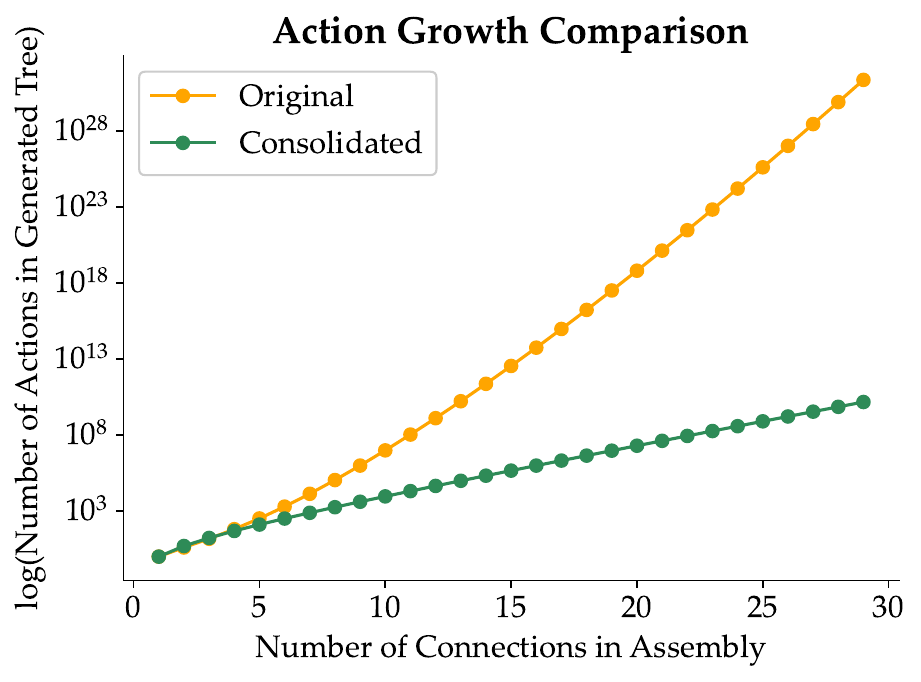}}
    
    \caption{The State and Action Growth Rate compared to the number of connections in the fully assembled structure. This directly translates to the growth rate of the nodes and edges in the tree graph $\mathcal{H}$, respectively, compared to the number of edges in graph $\mathcal{G}$. Note that the y-axis of these plots are on a logarithmic scale.}\label{fig:GrowthComparison}
\end{figure}

As seen in Fig.~\ref{fig:GrowthComparison}, despite our consolidation efforts on the state-action space, the rate of growth is exponential, which motivated our expansion of our sequential decision making framework to a Deep Q Network (DQN) approach.

\subsection*{Forward Assembly Approach}\label{sec:ForwardAssembly}
\begin{figure}
    \centering
    \includegraphics[width=0.95\columnwidth]{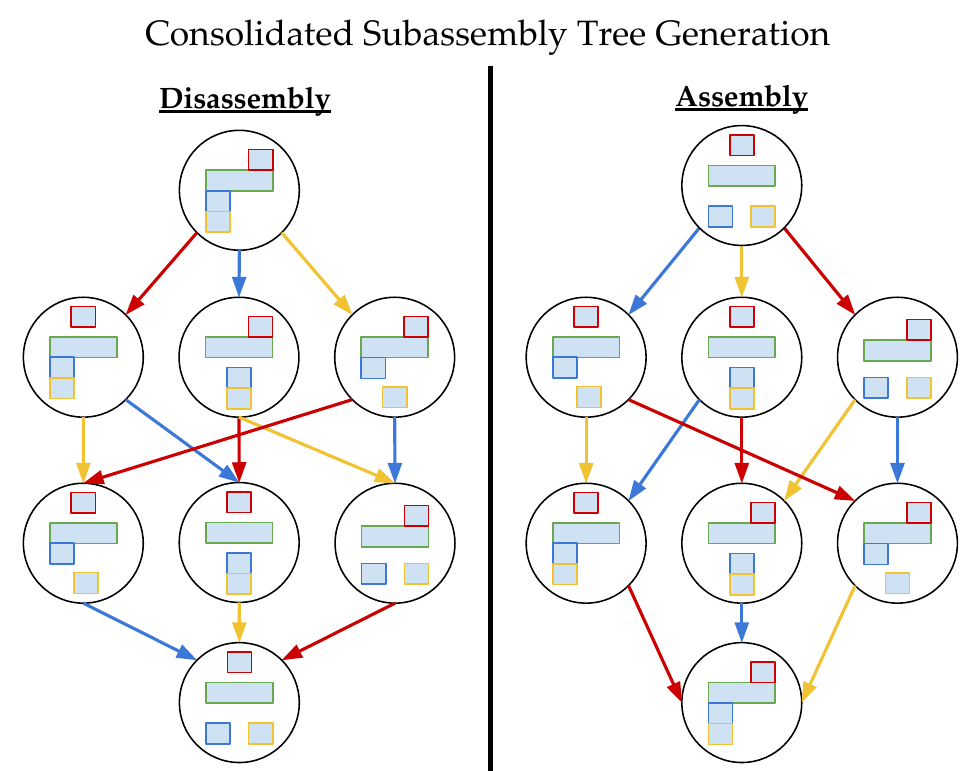}
      \caption{Showing that the consolidated tree graph structure is computationally symmetrical in structure for Assembly and Disassembly tasks.}
      \label{fig:consBoth}
\end{figure}

We primarily utilize Assembly-by-Disassembly to align our work with previous papers in the assembly field, as many cost functions and  general objective functions have already been engineered with this ordering in mind~\cite{RuzenaK.Bajcsy1989, Culbertson2019, Ghandi2015}. Additionally, a useful characteristic of our consolidated tree graph structure is that it is symmetrical. This means that our generated tree can be equivalently explored as a forward assembly graph if there are reward and cost functions that we can associate with these forward actions. Also notice that in this case, the fully disconnected assembly will be the initial state $s_i$ and thus the top node. Similarly, the action definition $a$ is adjusted to the attachment of a connection instead, and the state order in the consolidated tree graph is effectively flipped, as illustrated in Fig.~\ref{fig:consBoth}. Thus, with an appropriate forward objective function, our framework can also operate in the forward assembly direction with no major modifications to any of our proposed algorithms. In some cases, the costs of the forward assembly and backward disassembly tasks are the same, meaning that no change in weights is necessary, and our methods will continue to operate effectively. Most of the heuristics discussed in our Related Works Section~\ref{sec:relatedWorks} fail to adapt to a forward assembly approach.

\subsection*{Constraint Example}\label{sec:ConstraintExample}
\begin{figure}[ht]
    \centering
    \subfloat[Results with No Sequential Constraints]{\includegraphics[width=0.85\columnwidth]{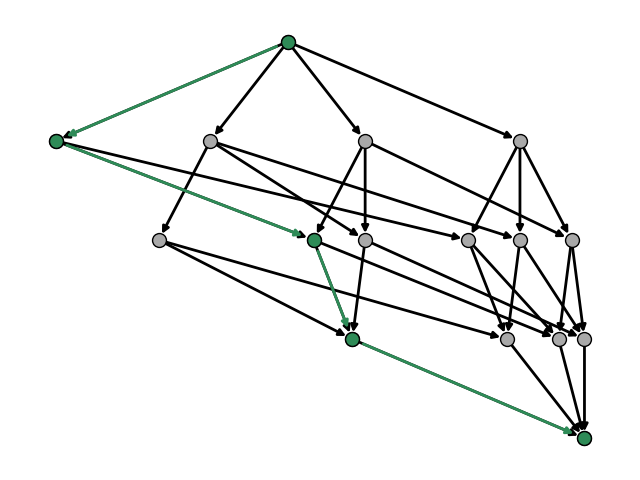}}
    
    \subfloat[Results with Sequential Constraints]{\includegraphics[width=0.85\columnwidth]{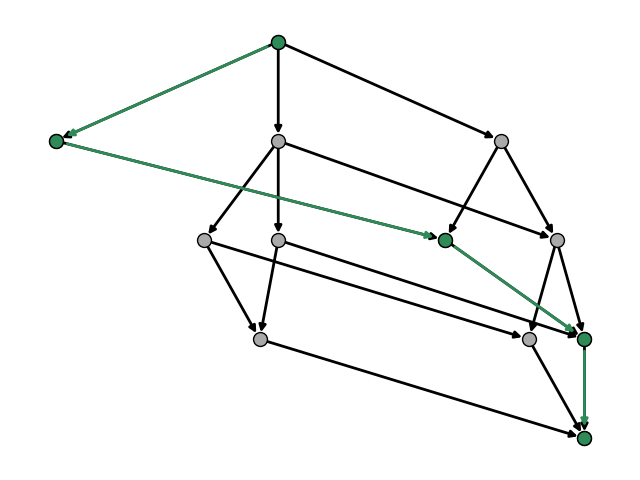}}
    \caption{The effect of adding a sequential constraint onto the consolidated state-action space $\mathcal{H}$ for a simple 4-Piece structure. A sequential constraint where the connection $(2,3)$ must be removed before removing $(1,2)$. Observe that this results in not only a difference in the size and shape of the state-action space, but also on the optimal route through it (shown in green).}
    \label{fig:cnstrnt}
\end{figure}

We have provided Fig.~\ref{fig:cnstrnt} to show the effects of adding a sequential constraint to this robotic assembly planning problem. Fig.~\ref{fig:cnstrnt} presents a visual representation of the before-and-after scenario. In the "before" segment, we showcase the conventional approach to robotic assembly planning, where no constraints are specified. In the "after" segment of the figure, we portray the effect of adding the sequential constraint where the connection (2, 3) must be removed
before removing (1, 2) in the assembly planning process. This constraint fundamentally changes the results by pruning branches from the disassembly tree $\mathcal{H}$ which allow for violation of this constraint. This results in not only a different shape for the disassembly tree, but also a reduction in the number of states and actions in the overall system. As a result our algorithms can better explore this reduced solution space to find the optimal assembly sequence, which can be different from the original strategy, as further illustrated in the figure.

Through this visual representation, we aim to clearly convey the advantages of our sequential constraint formulation. The figure showcases how the integration of this constraint optimizes the search process, thereby reducing runtimes and enhancing re planning capabilities, which ultimately leads to more efficient robotic assembly.

\subsection*{Graph-Based Value Iteration Algorithm}\label{sec:GraphValueAlgo}
As discussed in the body of the paper, we utilized Eqn.~\eqref{eq:value} to adapt the traditional value iteration algorithm to the robotic manufacturing domain, producing the following Graph-Based Value Iteration Algorithm~\ref{alg:graph_value_iteration}.
\begin{algorithm}
    \caption{Graph-Based Value Iteration Algorithm}
    \label{alg:graph_value_iteration}
    \begin{algorithmic}[1]
        \Require Graph $\mathcal{H}$
        \Require $\varepsilon > 0$  (Convergence Condition), $M$ (Maximum Number of Iterations)
        \Ensure $\pi$ (Deterministic Policy) s.t. $\pi \approx \pi^*$
        \Ensure Value Function $V(s)$
        \State Initialize $V_0(s)$ for all $s \in \mathcal{S}$ randomly (or some initial value based on a prior)
        \For{$k=0,1,2,...,N$}
        \State $\Delta \gets 0$
        \State $s_0 \gets s_i$\Comment{Predefined Initial State}
        \For{all $s \in \mathcal{H}$}
            \State $V_{k+1}(s) \gets \max_{a \in \mathcal{H}[s]} \left [R(s,a) + V_k(\mathcal{H}[s][a]) \right ]$\Comment{The next state $s'$ is deterministic}
            \State $\Delta \gets \max \{\Delta, |V_{k+1}(s) - V_k(s)|\}$
        \EndFor
        \If{$\Delta < \varepsilon$}
        \State \textbf{break}
        \EndIf
        \EndFor
        \State $V(s) \gets V_k(s)$ for all $s \in \mathcal{S}$
        \State $\pi \text{ s.t. } \pi(a|s)=\argmax_{a} \left[R(s,a)+V(\mathcal{H}[s][a])\right]$
        \State \Return $V(s), \pi(a|s)$
    \end{algorithmic}
\end{algorithm}

\subsection*{Epsilon-Greedy Policy}\label{sec:epsGreedy}

A $\varepsilon$-greedy policy $\pi$ is defined as follows:
\begin{equation} \label{eq:greedy}
    \pi(a \mid \hat{s})= \begin{cases}
    \underset{a}{\argmax} Q(\hat{s}, a) & \text { with probability } 1-\varepsilon \\ 
    a \sim \mathcal{U}\left(\mathcal{A}_{\hat{s}}\right) & \text { with probability }\varepsilon
    \end{cases}
\end{equation}
where $\mathcal{U}$ denotes a uniform sampling and $\mathcal{A}_{\hat{s}}$ is the action space available at the given state $\hat{s}$, which in the robotic manufacturing setting translates to the edges remaining in the subassembly $\hat{s}$.

\subsection*{Additional Results}\label{sec:AdditionalResults}
\begin{figure}
    \centering
    
    \subfloat[Resulting Path 1]{\includegraphics[width=0.95\columnwidth]{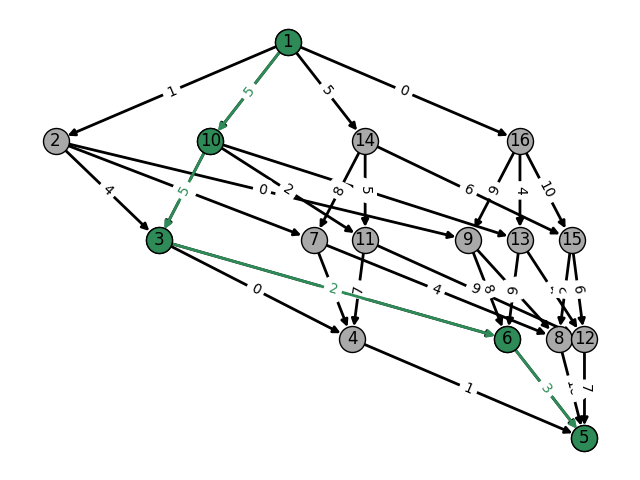}}
    
    \subfloat[Resulting Path 2]{\includegraphics[width=0.95\columnwidth]{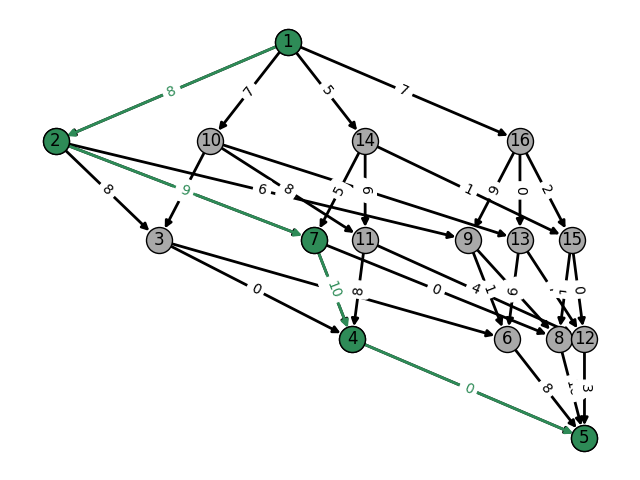}}
    
    \caption{Two examples of optimal paths through $\mathcal{H}$ with different choices of Reward functions. This form of $\mathcal{H}$ is specific to the deconstruction of the 4-Part, 4-Connection structure ``4Piece". For simplicity, all rewards are integers and the graph consolidation technique is utilized. As a reminder, Node 1 (i.e. top of the tree) corresponds to the fully assembled structure and Node 5 (i.e. the bottom of the tree) corresponds to the fully deconstructed state, as illustrated in~\ref{fig:treeGen2}}.
    \label{fig:opt_path}
\end{figure}

Using the value iteration method illustrated in Algorithm~\ref{alg:graph_value_iteration}, we find an optimal path from the start of the tree $\mathcal{H}$ to the end. This path indirectly prescribes a disassembly ordering, which can then be reversed to produce the optimal assembly sequence. Fig.~\ref{fig:opt_path} shows the optimal path for the disassembly of the simple example ``4Brick" (which is illustrated in Fig.~\ref{fig:Ex}). Additionally, Fig.~\ref{fig:opt_path} provides a simple example of how utilizing two different reward functions can change the optimal path through the state-action space, as well as the amount of reward acquired during the disassembly process.

\begin{figure}
    \centering
    \subfloat[``4Brick"]{\includegraphics[width=0.45\columnwidth]{figs/4Brick.pdf}}
    \subfloat[``4Brick" Reward Distribution]{\includegraphics[width=0.45\columnwidth]{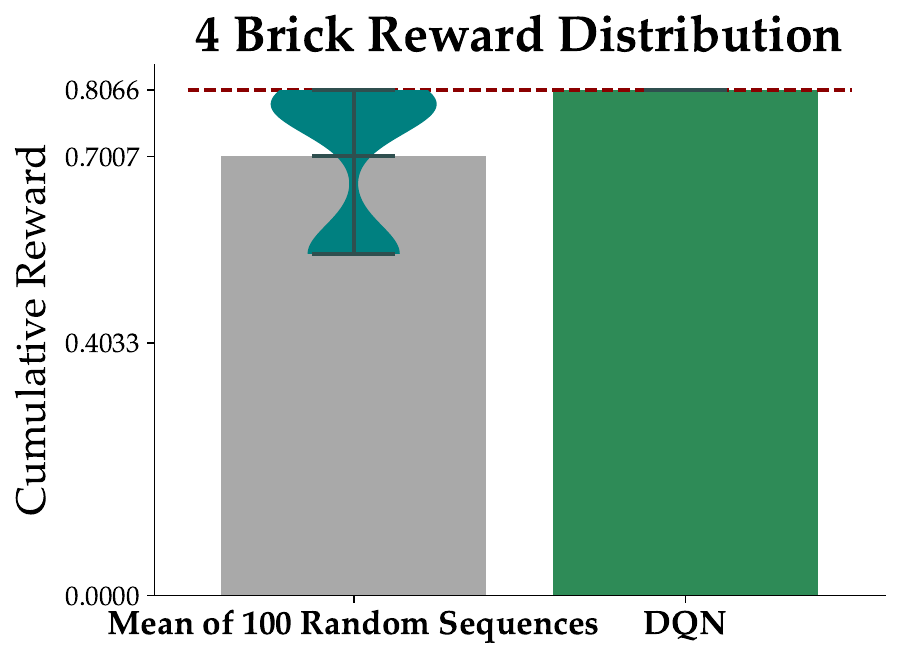}}\hfill
    
    \subfloat[``2x3"]{\includegraphics[width=0.45\columnwidth]{figs/2x3-Struct.pdf}}
    \subfloat[``2x3"]{\includegraphics[width=0.45\columnwidth]{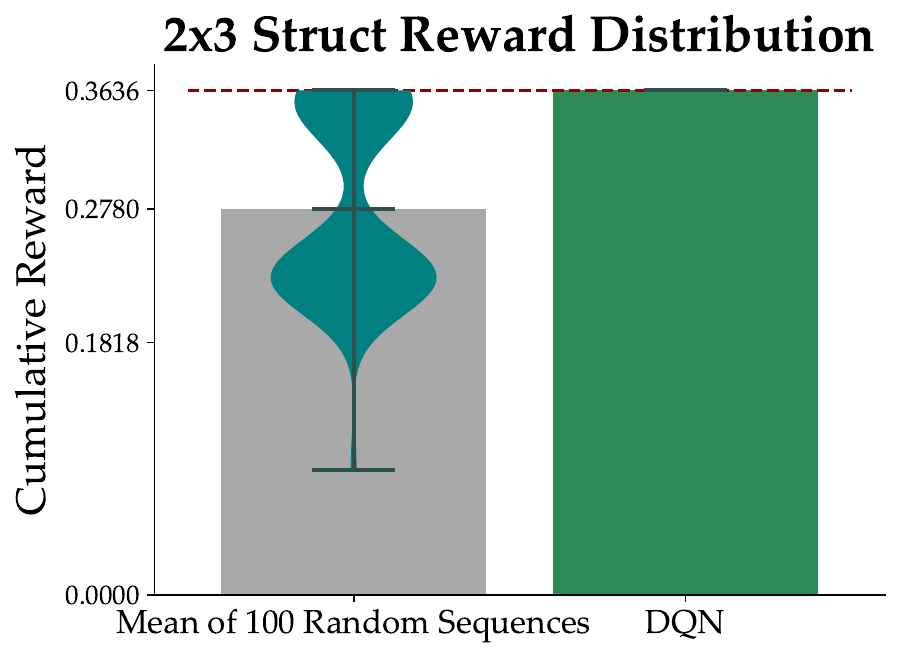}}\hfill

    \subfloat[``Lattice"]{\includegraphics[width=0.45\columnwidth]{figs/Lattice.pdf}}
    \subfloat[``Lattice"]{\includegraphics[width=0.45\columnwidth]{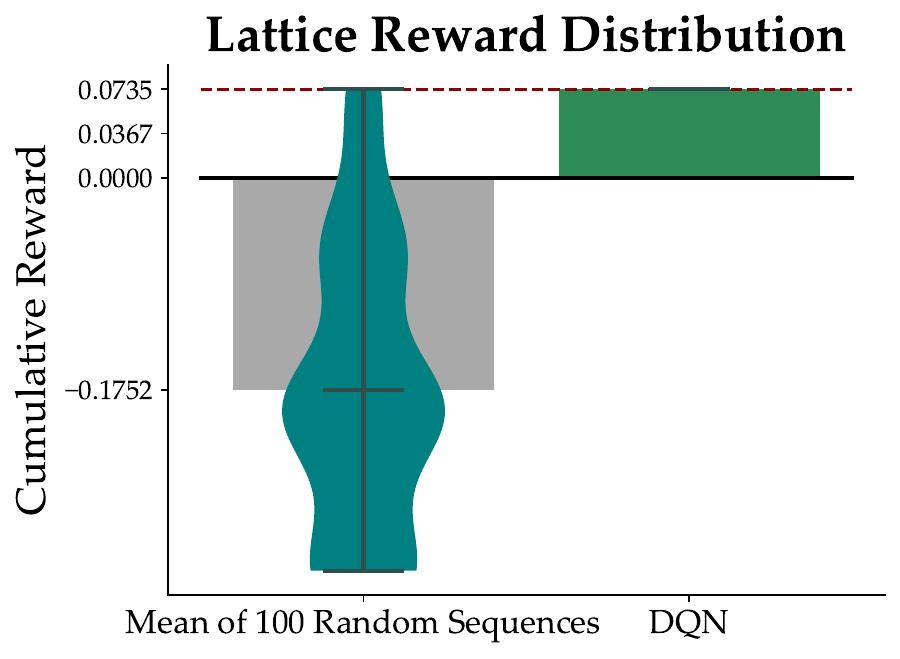}}
    
    \caption{The DQN results for the smaller structures. Our DQN approach matches or beats our full distribution of random sequences. As these scenarios are very small, there is high likelihood of some of the 100 random sequences recovering the optimal disassembly pattern.}\label{fig:SmallStructsDQN}
\end{figure}